%% file: main.tex
\documentclass{article}
\usepackage[preprint]{log_2024}			

\usepackage{booktabs}						
\usepackage{multirow}						
\usepackage{amsfonts}						
\usepackage{graphicx}						
\usepackage{duckuments}						

\usepackage[numbers,compress,sort]{natbib}	

\input{own_command}

\title[
Exploring Graph Structure Comprehension Ability of Multimodal Large Language Models: Case Studies
]
{
Exploring Graph Structure Comprehension Ability of Multimodal Large Language Models: Case Studies
}

\author[Z. Zhong and D. Mottin]{%
Zhiqiang Zhong \\
Aarhus University, Denmark \\
\email{zzhong@cs.au.dk}
\And
Davide Mottin \\
Aarhus University, Denmark \\
\email{davide@cs.au.dk}
}

\begin{document}

\maketitle

\begin{abstract} 
\input{pages/abstract}
\end{abstract}

\section{Introduction} 
\label{sec:introduction}
\input{pages/introduction}


\section{Exploring Graph Structure Comprehension Ability of Multimodal LLMs} 
\label{sec:methodology}
\input{pages/methodology}

\section{Results and Discussions} 
\label{sec:experiments}
\input{pages/experiments}

\section{Conclusion} 
\label{sec:conclusion}
\input{pages/conclusion}


\bibliographystyle{unsrtnat}
\bibliography{full_format_references}

\newpage
\appendix
\input{pages/appendix}

\end{document}

%% file: own_command.tex
\usepackage{adjustbox}
\usepackage{makecell}
\usepackage{multirow}
\usepackage[ruled,vlined,linesnumbered]{algorithm2e}
\usepackage{enumitem}
\usepackage{amsmath,amsthm}
\usepackage{xcolor,colortbl}
\usepackage{rotating}
\usepackage{mdframed}
\newmdtheoremenv[%
  backgroundcolor=gray!20,
  linecolor=red!60!black,
  linewidth=2pt,
  topline=false,
  rightline=false,
  skipabove=10pt,
  skipbelow=10pt,
  leftline=false]{regbox}{Box}

\newcommand{\specialcell}[2][c]{%
    \begin{tabular}[#1]{@{}c@{}}#2\end{tabular}
}

\newcommand{\eg}{\emph{e.g.}}

\newcommand{\model}{\textsc{GaI}\xspace}


%% file: pages/abstract.tex
Large Language Models (LLMs) have shown remarkable capabilities in processing various data structures, including graphs. 
While previous research has focused on developing textual encoding methods for graph representation, the emergence of multimodal LLMs presents a new frontier for graph comprehension. 
These advanced models, capable of processing both text and images, offer potential improvements in graph understanding by incorporating visual representations alongside traditional textual data.
This study investigates the impact of graph visualisations on LLM performance across a range of benchmark tasks at node, edge, and graph levels. 
Our experiments compare the effectiveness of multimodal approaches against purely textual graph representations. 
The results provide valuable insights into both the potential and limitations of leveraging visual graph modalities to enhance LLMs' graph structure comprehension abilities.

%% file: pages/introduction.tex
\input{figures/fig-architecture}

Recently, Large Language Models (LLMs) have revolutionised natural language processing and have been increasingly applied to diverse tasks beyond text generation and comprehension~\cite{PALM2,GPT4}. 
One area of growing interest is the application of LLMs to \emph{graph}-structured data, which is prevalent in numerous domains, \eg, social network analysis and bioinformatics~\cite{ZZM242,CMLJ23,ZZM24}. 

Conventionally, researchers have focused on developing textual encoding functions to represent graphs in a format digestible by LLMs~\cite{FHP24,CMLJ23,PFZTKRH24}. 
These methods have shown promise, enabling LLMs to perform various graph-related tasks with increasing accuracy. 
While this approach has shown promise, it faces inherent limitations in capturing the full complexity of graph structures, particularly in preserving spatial relationships and global structural properties~\cite{FHP24}.

The recent emergence of multimodal LLMs marks a significant milestone in AI development~\cite{GPT4,PALM2}. 
These advanced models, capable of processing both textual and visual information, open new avenues for enhancing machine comprehension of complex data structures. 
In the context of graph structure comprehension, this multimodal capability presents an exciting opportunity: \emph{the potential to leverage visual representations of graphs alongside their textual descriptions}.

This research aims to explore the potential of multimodal LLMs in graph comprehension tasks. 
We hypothesise that by leveraging both textual and visual representations of graphs, these models can achieve superior performance compared to their text-only representations. 
Our study focuses on a comprehensive set of benchmark tasks at the node, edge, and graph levels, providing a multifaceted evaluation of multimodal approaches in graph analysis.
Particularly, based on the designed framework as shown in Figure~\ref{fig:architecture}, we seek to address two research questions:
\emph{(i)} How does incorporating visual graph representations affect LLM performance on various graph-related tasks compared to purely textual representations?
\emph{(ii)} What are the limitations of current multimodal LLMs in processing graph visualisations, and how might these be addressed in future research?

%% file: figures/fig-architecture.tex
\begin{figure}[!ht]
\centering
\includegraphics[width=.9\linewidth]{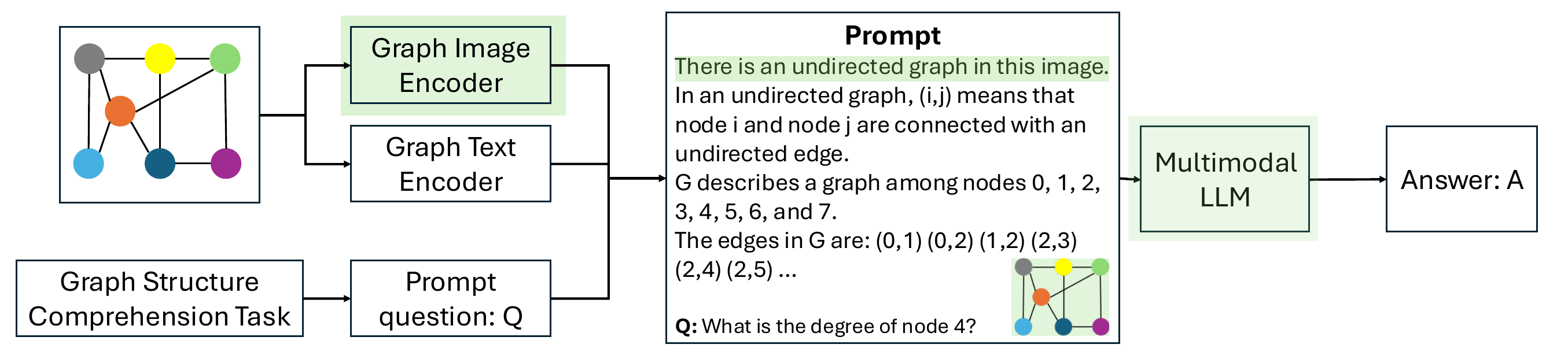}
\caption{
Overview of our framework ($\model^{+}$) for graph structure comprehension using multimodal LLMs. 
The newly added components, compared to \cite{FHP24}, are highlighted in \colorbox{green!14}{green} for clarity.
}
\label{fig:architecture}
\vspace{-3mm}
\end{figure}

%% file: pages/methodology.tex
Our empirical studies follow the GraphQA benchmark settings~\cite{FHP24}. 
Figure~\ref{fig:architecture} provides an overview of our framework, comprehending \underline{G}raph \underline{a}s \underline{I}mage ($\model^+$). 
Its simplified version, \model, indicates the only graph vision modality is included. 
We detail each component of our methodology below.

\textbf{Graph Generation.}
To systematically evaluate the graph comprehension capabilities of multimodal LLMs, we generated a diverse set of graphs using the Erdős–Rényi (ER) model~\cite{ER59}, following the approach of \citet{FHP24}. 
Our dataset comprises $500$ graphs, each containing between $5$ and $20$ nodes. 
This range allows us to assess the models' performance across varying graph complexities. 
Figure~\ref{fig:example} illustrates two example graphs from our dataset.

\textbf{Graph Text Encoder.}
While \citet{FHP24} propose several text encoding functions to represent graphs, we focused on two specific methods: adjacency and incident encoding. 
This choice was motivated by the need to visualise graphs as images, where complicated textual representations might be challenging to depict within a constrained visual space. 
These encoding methods provide a balance between informational content and visual clarity.

\textbf{Graph Visualiser.}
The graph visualiser component generates visual representations of the structural graphs. 
While there can be numerous variations in visual aspects such as background colours, layouts, and node shapes, we opted for a standardised approach using Matplotlib~\cite{matplotlib09} with default settings. 
This decision ensures consistency across our visual graph representations. 
All graphs are plotted to a fixed size to maintain uniformity. We acknowledge that different visualisation techniques could influence results, and we identify this as an area for future investigation.

\textbf{Prompt Construction.}
We adopted all prompt designs from \cite{FHP24}, which include: Zero-shot prompting (\textsc{ZERO-SHOT}), Few-shot in-context learning (\textsc{FEW-SHOT}), Chain-of-thought (\textsc{CoT}), Zero-shot CoT prompting (\textsc{ZERO-COT}) and Bag prompting (\textsc{COT-BAG}). 
For scenarios where a visual graph representation is available, we augmented the prompts by prepending the sentence: "There is an undirected graph in this image." 
This modification ensures that the LLM is aware of the presence of visual information.
Our study encompasses a comprehensive set of graph structure comprehension tasks, including: Node tasks: node degree, connected nodes; Edge tasks: edge existence, shortest path; and Graph tasks: node count, edge count, cycle check, triangle counting. 
This diverse set of tasks allows us to evaluate the models' performance across various aspects of graph comprehension.

\textbf{LLMs.}
Our study focuses on LLMs in a black-box setup, where the model parameters are fixed, and the system only consumes and produces text. 
This setting reflects the most common scenario for practical LLM usage. 
We selected two state-of-the-art multimodal LLMs for our experiments: GPT-4~\cite{GPT4}, GPT-4o~\cite{GPT4}. 
These models represent the current pinnacle of multimodal language models, capable of processing both text and image inputs.

%% file: pages/experiments.tex
Our experimental results are summarised in Tables~\ref{table:results_gpt4o} and \ref{table:results_gpt4}. 
We discuss our findings in detail below:

\input{tables/table-results-er-4o}
\input{tables/table-results-er-4}

\textbf{Superior performance of multimodal LLMs.}
A impressive observation from our results 
is the markedly superior performance of GPT-4o and GPT-4 compared to the PaLM model. 
In several tasks, these newer models demonstrate near-perfect accuracy, correctly answering questions about graph structures for almost all test cases. 
This substantial improvement indicates that recent advancements in multimodal LLMs have significantly enhanced their graph structure comprehension abilities.

\textbf{Impact of graph visualisation.}
Our results show that incorporating graph visualisations can enhance LLMs' graph comprehension, though this effect is not uniform across all tasks. 
The impact of visual input varies depending on: \emph{(i)} The complexity of the graph structure. 
\emph{(ii)} The specific nature of the task (\eg, local \emph{vs.} global graph properties). 
For instance, tasks involving global properties (\eg, cycle detection) seem to benefit more from visual input compared to local tasks (\eg, node degree).

\textbf{Limitations of visual-only input.}
Interestingly, we found that providing only graph visualisations, without accompanying textual descriptions, is insufficient for LLMs to fully comprehend graph structures. 
This observation highlights the complementary nature of visual and textual information in graph comprehending tasks. 

\textbf{Comparison with specialised graph encoding models.}
Our comparison with the work of \cite{PFZTKRH24}, which uses neural networks to encode graph information for LLMs, reveals that our multimodal LLM approach outperforms these carefully trained models in graph structure comprehension tasks. 
This finding is significant because it suggests that:
\emph{(i)} General-purpose multimodal LLMs can compete with, and even surpass, specialised graph encoding models.
\emph{(ii)} The versatility of multimodal LLMs allows them to adapt effectively to graph comprehending tasks without task-specific training.



\input{tables/table-results-er-nn}


\textbf{Challenges in graph visualisation.}
Figure~\ref{fig:example} illustrates two contrasting examples of graph visualisation: a simple graph with clear visual representation and a complex graph where \model provides incorrect responses. 
This comparison highlights a critical challenge in our approach: the effective visualisation of graphs for multimodal LLMs.
The disparity in performance between simple and complex graphs raises several important questions:
\emph{(i)} How does graph complexity affect the model's ability to extract relevant information from visualisations?
\emph{(ii)} What are the optimal ways to visually represent different types of graph structures?
\emph{(iii)} How can we balance information density and visual clarity in graph representations?
These observations underscore the need for further research into graph visualisation techniques that are optimised for LLM comprehension. 
Future work should explore various visualisation strategies, potentially incorporating:
\emph{(i)} Sampling-based interactive or dynamic graph representations. 
\emph{(ii)} Hierarchical visualisations for complex graphs. 
\emph{(iii)} Novel encoding techniques that highlight relevant graph properties. 

%% file: tables/table-results-er-4o.tex
\begin{table}[!ht]
\centering
\resizebox{1.\columnwidth}{!}{%
\begin{tabular}{c|c|cccccccc}
    \hline
    \hline
    \textbf{Prompt} & \textbf{Encoding} & \specialcell{\textbf{Edge} \\ \textbf{Existence}} & \specialcell{\textbf{Node} \\ \textbf{degree}} & \specialcell{\textbf{Node} \\ \textbf{count}} & \specialcell{\textbf{Edge} \\ \textbf{count}} & \specialcell{\textbf{Connected} \\ \textbf{nodes}} & \specialcell{\textbf{Cycle} \\ \textbf{check}} & \specialcell{\textbf{Triangle} \\ \textbf{counting}} & \specialcell{\textbf{Shortest} \\ \textbf{path}} \\ 
    \hline
    \multirow{7}{*}{\begin{sideways}{\normalsize \textsc{zero-shot}}\end{sideways}}
    & GraphQA~\cite{FHP24} & 49.0* & 25.0* & 24.2* & 15.0* & 53.8* & 82.0* & 1.5* & 11.5* \\
    \cline{2-10}
    & Adjacency & 96.2 & 75.8 & \textbf{100.0} & 67.4 & 76.8 & 96.2 & 33.0 & 69.0 \\
    & $\model_{\textsc{Adj}}$ & 74.6 & 55.8 & 93.4 & 21.2 & 35.8 & \cellcolor{gray!25} 97.0 & 26.4 & 53.0 \\
    & $\model_{\textsc{Adj}}^{+}$ & 96.4 & 70.8 & \textbf{100.0} & 65.8 & 76.4 & \cellcolor{blue!25} \textbf{98.8} & 30.6 & 63.2 \\
    \cline{2-10}
    & Incident & 97.0 & 84.4 & \textbf{100.0} & 54.2 & 89.4 & 93.8 & 26.2 & 68.6 \\
    & $\model_{\textsc{Inc}}$ & 77.2 & 50.8 & 92.2 & 22.2 & 36.6 & \cellcolor{gray!25} 96.6 & 25.2 & 56.8 \\
    & $\model_{\textsc{Inc}}^{+}$ & \cellcolor{blue!25} 99.2 & 78.6 & \textbf{100.0} & \cellcolor{blue!25} 55.4 & 88.0 & \cellcolor{blue!25} 98.4 & \cellcolor{blue!25} 26.6 & \cellcolor{blue!25} 69.2 \\
    \hline
    \multirow{7}{*}{\begin{sideways}{\normalsize \textsc{zero-cot}}\end{sideways}} 
    & GraphQA~\cite{FHP24} & 41.4* & 26.6* & 19.4* & 12.2* & 35.2* & 46.2* & 12.7* & 33.6* \\
    \cline{2-10}
    & Adjacency & 92.2 & 76.6 & 95.4 & 73.4 & 82.6 & 96.6 & \textbf{33.8} & 71.6 \\
    & $\model_{\textsc{Adj}}$ & 60.2 & 46.4 & 94.8 & 24.2 & 35.4 & \cellcolor{gray!25} 97.2 & 25.4 & 52.4 \\
    & $\model_{\textsc{Adj}}^{+}$ & 90.8 & 69.0 & \cellcolor{blue!25} 99.6 & 66.8 & 79.2 & \cellcolor{blue!25} 98.4 & 31.4 & 70.8 \\
    \cline{2-10}
    & Incident & 97.2 & 72.6 & 97.6 & 54.6 & 89.2 & 90.6 & 28.6 & 73.6 \\
    & $\model_{\textsc{Inc}}$ & 62.2 & 47.6 & 93.8 & 24.8 & 35.0 & \cellcolor{gray!25} 96.2 & 24.6 & 52.8 \\
    & $\model_{\textsc{Inc}}^{+}$ & \cellcolor{blue!25} 98.2 & 72.6 & \cellcolor{blue!25} \textbf{100.0} & \cellcolor{blue!25} 62.0 & 86.6 & \cellcolor{blue!25} 97.0 & 26.0 & \cellcolor{blue!25} 74.2 \\
    \hline
    \multirow{7}{*}{\begin{sideways}{\normalsize \textsc{few-shot}}\end{sideways}} 
    & GraphQA~\cite{FHP24} & 42.8* & 33.6* & 51.2* & 18.6* & 36.6* & 47.8* & 3.0* & 22.7* \\
    \cline{2-10}
    & Adjacency & 93.0 & 69.6 & \textbf{100.0} & 67.8 & 82.4 & 93.2 & 29.2 & 67.8 \\
    & $\model_{\textsc{Adj}}$ & 84.0 & 49.4 & 94.0 & 19.6 & 32.0 & \cellcolor{gray!25} 96.8 & 24.8 & 62.4 \\
    & $\model_{\textsc{Adj}}^{+}$ & \cellcolor{blue!25} 96.4 & \cellcolor{blue!25} 70.0 & 99.4 & 64.4 & 80.6 & \cellcolor{blue!25} 94.6 & 27.0 & \cellcolor{blue!25} 68.6 \\
    \cline{2-10}
    & Incident & \textbf{99.4} & \textbf{94.0} & \textbf{100.0} & 30.2 & 90.4 & 94.2 & 24.0 & 75.2 \\
    & $\model_{\textsc{Inc}}$ & 83.4 & 49.6 & 92.8 & 20.2 & 34.2 & \cellcolor{gray!25} 96.4 & \cellcolor{gray!25} 24.4 & 60.0 \\
    & $\model_{\textsc{Inc}}^{+}$ & 98.6 & 90.8 & 98.2 & \cellcolor{blue!25} 48.0 & 89.4 & \cellcolor{blue!25} 96.6 & \cellcolor{blue!25} 27.0 & \cellcolor{blue!25} 75.6 \\
    \hline
    \multirow{7}{*}{\begin{sideways}{\normalsize \textsc{cot}}\end{sideways}}
    & GraphQA~\cite{FHP24} & 46.6* & 75.0* & 57.6* & 25.2* & 30.2* & 62.6* & 8.1* & 38.6* \\
    \cline{2-10}
    & Adjacency & 92.2 & 70.2 & \textbf{100.0} & 67.8 & 84.8 & 93.4 & 28.6 & 70.0 \\
    & $\model_{\textsc{Adj}}$ & 84.2 & 47.4 & 92.6 & 16.2 & 30.8 & \cellcolor{gray!25} 96.6 & 24.6 & 61.4 \\
    & $\model_{\textsc{Adj}}^{+}$ & \cellcolor{blue!25} 95.0 & \cellcolor{blue!25} 71.8 & 99.8 & 63.8 & 80.4 & \cellcolor{blue!25} 95.8 & 27.4 & 69.0 \\
    \cline{2-10}
    & Incident & 98.4 & 92.2 & 99.8 & 27.0 & \textbf{90.2} & 95.4 & 24.4 & \textbf{76.4} \\
    & $\model_{\textsc{Inc}}$ & 84.4 & 48.4 & 94.0 & 18.8 & 31.8 & \cellcolor{gray!25} 97.0 & 24.2 & 60.6 \\
    & $\model_{\textsc{Inc}}^{+}$ & 98.4 & 89.8 & 98.8 & \cellcolor{blue!25} 36.0 & 89.2 & \cellcolor{blue!25} 97.2 & \cellcolor{blue!25} 25.6 & 74.8 \\
    \hline
    \multirow{7}{*}{\begin{sideways}{\normalsize \textsc{cot-bag}}\end{sideways}} 
    & GraphQA~\cite{FHP24} & 45.8* & 75.2* & 51.2* & 25.0* & 41.0* & 63.0* & 8.1* & 40.4* \\
    \cline{2-10}
    & Adjacency & 94.0 & 71.2 & \textbf{100.0} & \textbf{70.4} & 83.6 & 92.6 & 27.0 & 68.2 \\
    & $\model_{\textsc{Adj}}$ & 86.6 & 48.8 & 93.4 & 17.6 & 31.0 & \cellcolor{gray!25} 96.6 & 25.4 & 60.6 \\
    & $\model_{\textsc{Adj}}^{+}$ & \cellcolor{blue!25} 96.0 & 66.0 & 99.8 & 65.6 & 79.6 & \cellcolor{blue!25} 93.6 & 27.0 & 67.6 \\
    \cline{2-10}
    & Incident & 98.8 & 90.6 & 99.8 & 22.0 & \textbf{90.2} & 93.4 & 24.2 & 74.2 \\
    & $\model_{\textsc{Inc}}$ & 83.8 & 49.6 & 93.4 & 17.0 & 31.8 & \cellcolor{gray!25} 97.0 & 24.2 & 60.4 \\
    & $\model_{\textsc{Inc}}^{+}$ & \cellcolor{blue!25} 99.0 & 90.2 & 98.8 & \cellcolor{blue!25} 23.0 & 89.0 & \cellcolor{blue!25} 95.6 & 23.6 & \cellcolor{blue!25} 75.4 \\
    \hline
    \hline
\end{tabular}
}
\caption{
Comparison of various graph encoder functions based on their accuracy on different graph tasks using GPT-40. 
* indicates the results reported in \cite{FHP24} based on PaLM~\cite{PALM2}. 
The results where $\model^{+}$ makes improvements are highlighted in \colorbox{blue!25}{blue}. 
The results where $\model$ outperforms the corresponding baseline are highlighted in \colorbox{gray!25}{gray}. 
}
\vspace{-6mm}
\label{table:results_gpt4o}
\end{table}

%% file: tables/table-results-er-4.tex
\begin{table}[!ht]
\centering
\resizebox{1.\columnwidth}{!}{%
\begin{tabular}{c|c|cccccccc}
    \hline
    \hline
    \textbf{Prompt} & \textbf{Encoding} & \specialcell{\textbf{Edge} \\ \textbf{Existence}} & \specialcell{\textbf{Node} \\ \textbf{degree}} & \specialcell{\textbf{Node} \\ \textbf{count}} & \specialcell{\textbf{Edge} \\ \textbf{count}} & \specialcell{\textbf{Connected} \\ \textbf{nodes}} & \specialcell{\textbf{Cycle} \\ \textbf{check}} & \specialcell{\textbf{Triangle} \\ \textbf{counting}} & \specialcell{\textbf{Shortest} \\ \textbf{path}} \\ 
    \hline
    \multirow{7}{*}{\begin{sideways}{\normalsize \textsc{zero-shot}}\end{sideways}}
    & GraphQA~\cite{FHP24} & 49.0* & 25.0* & 24.2* & 15.0* & 53.8* & 82.0* & 1.5* & 11.5* \\
    \cline{2-10}
    & Adjacency & 94.2 & 44.2 & 99.4 & 63.2 & 74.8 & 96.0 & 23.6 & 74.4 \\
    & $\model_{\textsc{Adj}}$ & 72.4 & 43.2 & 82.2 & 20.4 & 27.6 & 95.2 & 23.6 & 50.0 \\
    & $\model_{\textsc{Adj}}^{+}$ & 92.0 & \cellcolor{blue!25} 70.0 & \cellcolor{blue!25} \textbf{100.0} & 61.4 & 74.4 & \cellcolor{blue!25} \textbf{98.6} & 27.8 & 55.6 \\
    \cline{2-10}
    & Incident & 97.6 & 64.8 & 99.2 & 42.6 & 89.2 & 88.4 & 26.4 & 76.8 \\
    & $\model_{\textsc{Inc}}$ & 74.8 & 43.8 & 81.6 & 20.4 & 28.0 & \cellcolor{gray!25} 95.4 & 22.8 & 50.0 \\
    & $\model_{\textsc{Inc}}^{+}$ & 95.6 & \cellcolor{blue!25} 66.4 & \cellcolor{blue!25} 99.8 & \cellcolor{blue!25} 48.2 & \cellcolor{blue!25} 89.6 & \cellcolor{blue!25} 97.6 & \cellcolor{blue!25} 27.0 & 62.0 \\
    \hline
    \multirow{7}{*}{\begin{sideways}{\normalsize \textsc{zero-cot}}\end{sideways}} 
    & GraphQA~\cite{FHP24} & 41.4* & 26.6* & 19.4* & 12.2* & 35.2* & 46.2* & 12.7* & 33.6* \\
    \cline{2-10}
    & Adjacency & 95.0 & 61.6 & 99.4 & 63.8 & 77.0 & 96.0 & 31.0 & 71.4 \\
    & $\model_{\textsc{Adj}}$ & 73.4 & 40.8 & 79.0 & 19.8 & 26.8 & 95.6 & 23.4 & 50.8 \\
    & $\model_{\textsc{Adj}}^{+}$ & 83.6 & \cellcolor{blue!25} 67.6 & \cellcolor{blue!25} \textbf{100.0} & 59.2 & 73.6 & \cellcolor{blue!25} 97.8 & \cellcolor{blue!25} \textbf{31.2} & 58.8 \\
    \cline{2-10}
    & Incident & 98.0 & 76.2 & 99.6 & 39.6 & 88.8 & 88.4 & 26.8 & 76.8 \\
    & $\model_{\textsc{Inc}}$ & 74.4 & 42.6 & 76.4 & 19.0 & 24.8 & \cellcolor{gray!25} 96.0 & 21.8 & 49.8 \\
    & $\model_{\textsc{Inc}}^{+}$ & 93.0 & 71.4 & \cellcolor{blue!25} \textbf{100.0} & \cellcolor{blue!25} 50.2 & 88.6 & \cellcolor{blue!25} 97.4 & \cellcolor{blue!25} 27.8 & 63.2 \\
    \hline
    \multirow{7}{*}{\begin{sideways}{\normalsize \textsc{few-shot}}\end{sideways}} 
    & GraphQA~\cite{FHP24} & 42.8* & 33.6* & 51.2* & 18.6* & 36.6* & 47.8* & 3.0* & 22.7* \\
    \cline{2-10}
    & Adjacency & 95.6 & 63.2 & \textbf{100.0} & 60.6 & 75.0 & 94.6 & 25.6 & 69.0 \\
    & $\model_{\textsc{Adj}}$ & 80.4 & 45.6 & 85.0 & 20.0 & 25.2 & 93.0 & 22.6 & 61.8 \\
    & $\model_{\textsc{Adj}}^{+}$ & 94.4 & \cellcolor{blue!25} 65.4 & 99.6 & \cellcolor{blue!25} 61.8 & \cellcolor{blue!25} 76.8 & \cellcolor{blue!25} 95.0 & \cellcolor{blue!25} 29.0 & 65.4 \\
    \cline{2-10}
    & Incident & 98.2 & 92.6 & \textbf{100.0} & 24.4 & 90.6 & 90.4 & 27.6 & 74.4 \\
    & $\model_{\textsc{Inc}}$ & 80.2 & 45.0 & 82.4 & 21.6 & 26.2 & \cellcolor{gray!25} 92.6 & 23.2 & 61.0 \\
    & $\model_{\textsc{Inc}}^{+}$ & 97.0 & 91.2 & 98.6 & \cellcolor{blue!25} 30.0 & 90.4 & \cellcolor{blue!25} 94.6 & 27.6 & 69.8 \\
    \hline
    \multirow{7}{*}{\begin{sideways}{\normalsize \textsc{cot}}\end{sideways}}
    & GraphQA~\cite{FHP24} & 46.6* & 75.0* & 57.6* & 25.2* & 30.2* & 62.6* & 8.1* & 38.6* \\
    \cline{2-10}
    & Adjacency & 95.4 & 64.2 & 99.6 & 63.6 & 80.2 & 95.2 & 28.8 & 69.0 \\
    & $\model_{\textsc{Adj}}$ & 79.6 & 43.4 & 91.8 & 19.2 & 24.6 & 93.6 & 22.4 & 60.2 \\
    & $\model_{\textsc{Adj}}^{+}$ & 94.6 & \cellcolor{blue!25} 66.6 & \cellcolor{blue!25} 99.8 & 63.2 & 79.8 & 95.0 & \cellcolor{blue!25} 29.2 & 68.2 \\
    \cline{2-10}
    & Incident & \textbf{98.8} & \textbf{93.8} & 99.8 & 26.2 & 90.0 & 90.6 & 26.2 & 74.4 \\
    & $\model_{\textsc{Inc}}$ & 80.8 & 44.4 & 91.2 & 18.2 & 25.8 & \cellcolor{gray!25} 93.0 & 24.2 & 61.2 \\
    & $\model_{\textsc{Inc}}^{+}$ & 97.0 & 93.6 & 97.8 & \cellcolor{blue!25} 28.6 & \cellcolor{blue!25} 90.4 & \cellcolor{blue!25} 95.6 & 26.2 & 71.0 \\
    \hline
    \multirow{7}{*}{\begin{sideways}{\normalsize \textsc{cot-bag}}\end{sideways}} 
    & GraphQA~\cite{FHP24} & 45.8* & 75.2* & 51.2* & 25.0* & 41.0* & 63.0* & 8.1* & 40.4* \\
    \cline{2-10}
    & Adjacency & 95.8 & 63.0 & \textbf{100.0} & 63.8 & 81.8 & 96.0 & 27.6 & 69.0 \\
    & $\model_{\textsc{Adj}}$ & 78.8 & 42.4 & 90.8 & 18.8 & 25.2 & 92.6 & 24.2 & 59.6 \\
    & $\model_{\textsc{Adj}}^{+}$ & \cellcolor{blue!25} 96.0 & \cellcolor{blue!25} 66.0 & 99.4 & \cellcolor{blue!25} \textbf{64.0} & 80.8 & \cellcolor{blue!25} 96.6 & \cellcolor{blue!25} 29.2 & 68.2 \\
    \cline{2-10}
    & Incident & 98.0 & 93.2 & \textbf{100.0} & 24.4 & \textbf{90.8} & 89.8 & 25.6 & \textbf{76.2} \\
    & $\model_{\textsc{Inc}}$ & 81.0 & 42.8 & 91.4 & 19.6 & 23.4 & \cellcolor{gray!25} 92.8 & 23.6 & 61.6 \\
    & $\model_{\textsc{Inc}}^{+}$ & 97.4 & 92.0 & 98.8 & \cellcolor{blue!25} 27.2 & 90.0 & \cellcolor{blue!25} 95.6 & 25.6 & 69.4 \\
    \hline
    \hline
\end{tabular}
}
\caption{
Comparison of various graph encoder functions based on their accuracy on different graph tasks using GPT-4-turbo. 
* indicates the results reported in \cite{FHP24} based on PaLM~\cite{PALM2}. 
The results where $\model^{+}$ makes improvements are highlighted in \colorbox{blue!25}{blue}. 
The results where $\model$ outperforms the corresponding baseline are highlighted in \colorbox{gray!25}{gray}. 
}
\vspace{-6mm}
\label{table:results_gpt4}
\end{table}

%% file: tables/table-results-er-nn.tex
\begin{table}[!ht]
\centering
\resizebox{1.\columnwidth}{!}{%
\begin{tabular}{c|cccccccc}
    \hline
    \hline
    \textbf{Method} & \specialcell{\textbf{Edge} \\ \textbf{Existence}} & \specialcell{\textbf{Node} \\ \textbf{degree}} & \specialcell{\textbf{Node} \\ \textbf{count}} & \specialcell{\textbf{Edge} \\ \textbf{count}} & \specialcell{\textbf{Connected} \\ \textbf{nodes}} & \specialcell{\textbf{Cycle} \\ \textbf{check}} & \specialcell{\textbf{Triangle} \\ \textbf{counting}} & \specialcell{\textbf{Shortest} \\ \textbf{path}} \\ 
    \hline
    GraphQA~\cite{FHP24} & 49.0* & 75.2* & 57.6* & 25.2* & 53.8* & 82.0* & 12.7* & 40.4* \\
    GCN~\cite{KW17} & 68.0§ & 26.4§ & 74.6§ & 5.6§ & 26.4§ & 96.4§ & 20.8§ & 60.4§ \\
    GraphToken~\cite{PFZTKRH24} & 73.8§ & \textbf{96.2}§ & 99.6§ & 42.6§ & 26.4§ & 95.6§ & \textbf{34.8}§ & 63.8§ \\
    \model & \textbf{99.4} & 94.0 & \textbf{100.0} & \textbf{70.4} & \textbf{90.8} & \textbf{98.8} & 33.8 & \textbf{76.4} \\
    \hline
    \hline
\end{tabular}
}
\caption{
Comparison of various graph encoder functions based on their accuracy on different graph tasks. 
* indicates the best results reported in \cite{FHP24} based on PaLM~\cite{PALM2}. 
§ indicates the results reported in ~\cite{PFZTKRH24}.
The best performances are highlighted in \textbf{Bold}. 
}
\vspace{-6mm}
\label{table:results_nn}
\end{table}

%% file: pages/conclusion.tex
This study explored the graph structure comprehension abilities of multimodal LLMs through a series of empirical evaluations. 
We highlight the potential of multimodal LLMs for advancing graph structure comprehension tasks and suggests promising directions for future work in improving graph visualisations and multimodal integration.

%% file: pages/appendix.tex
\section{Appendix}

\input{figures/fig-example}

%% file: figures/fig-example.tex
\begin{figure}[!ht]
\centering
\includegraphics[width=.5\linewidth]{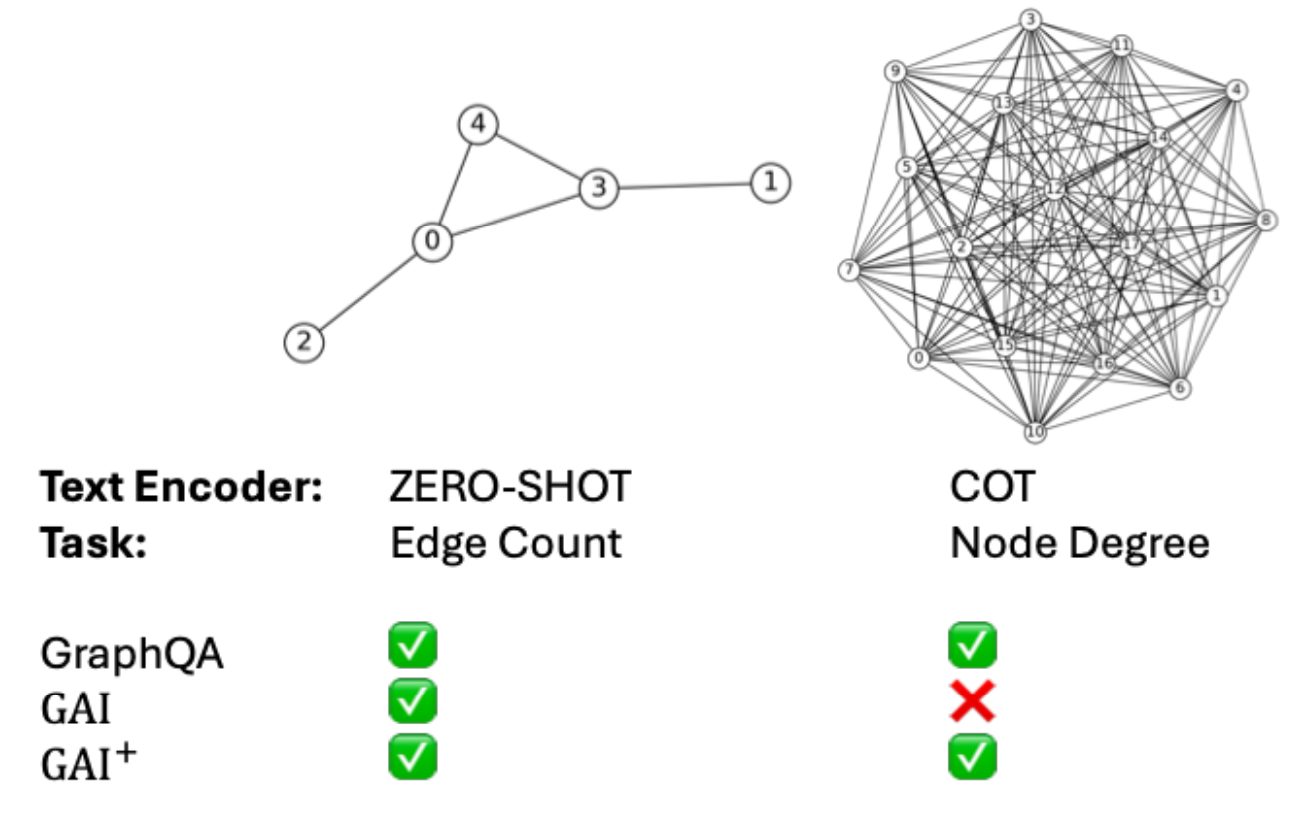}
\caption{
Illustrations of input images and the correctness of different models. 
}
\label{fig:example}
\end{figure}